\documentclass[10pt, conference, compsocconf]{IEEEtran}
\usepackage{amsthm}
\usepackage{amsmath}
\usepackage[noend]{algorithmic}
\usepackage{algorithm}
\usepackage{url}
\usepackage{subfigure}
\usepackage{multirow}
\usepackage{epsfig}
\usepackage{paralist}

\newcommand{\figDir}{}

\theoremstyle{definition}
\newtheorem{mydef}{Definition}

\begin{document}

%\title{Time Series Shapelets Classification - Nearly Best is Good Enough}
\title{Fast Randomized Model Generation for Shapelet-Based Time Series Classification}
%Another possibility: Time Series Shapelets Classification - Perfectionism is not Beneficial
\author{Daniel Gordon \qquad Danny Hendler \qquad Lior Rokach 
\\ Ben-Gurion University of The Negev 
\\Deutsche Telekom Laboratories
\\Be'er Sheva, Israel
\\gordonda@cs.bgu.ac.il,hendlerd@cs.bgu.ac.il,liork@bgu.ac.il}
%
%\numberofauthors{3}
%\author{
%% 1st. author
%\alignauthor
%Daniel Gordon\\
%       \affaddr{Ben-Gurion University of The Negev} \& \\
%       \affaddr{Deutsche Telekom Laboratories}\\
%       \affaddr{Be'er Sheva, Israel}\\
%       \email{gordonda@cs.bgu.ac.il}
%% 2nd. author
%\alignauthor
%Danny Hendler\\
%       \affaddr{Ben-Gurion University of The Negev} \& \\
%       \affaddr{Deutsche Telekom Laboratories}\\
%       \affaddr{Be'er Sheva, Israel}\\
%       \email{hendlerd@cs.bgu.ac.il}
%% 3rd. author
%\alignauthor Lior Rokach\\
%       \affaddr{Ben-Gurion University of The Negev} \& \\
%       \affaddr{Deutsche Telekom Laboratories}\\
%       \affaddr{Be'er Sheva, Israel}\\
%       \email{liork@bgu.ac.il}
%}

\maketitle
\begin{abstract}
Time series classification is a field which has drawn much attention
over the past decade.
A new approach for classification of time series uses classification trees based on shapelets.
A shapelet is a subsequence extracted from one of the time series in the dataset.
A disadvantage of this approach is the time required for building the shapelet-based classification
tree. The search for the best shapelet requires examining all
subsequences of all lengths from all time series in the training set.

%We introduce quasi-anytime algorithms for finding the best
%shapelets, allowing a user to bound the time required for
%building a shapelets-based classification tree.
%A user may
%ask a quasi-anytime algorithm at any point in time to re-
%turn a classification tree, in which case the algorithm builds
%and returns a classification tree model, based on the best
%shapelets found so far.

A key goal of this work was to find an evaluation order of the
shapelets space which enables fast convergence to an accurate model. The comparative analysis we conducted clearly indicates
that a random evaluation order yields the best results. Our empirical analysis of the distribution of high-quality shapelets within the shapelets space provides insights into why randomized shapelets sampling is superior to alternative evaluation orders.

We present an algorithm for randomized model generation for shapelet-based classification that converges extremely quickly to a model with surprisingly high accuracy after evaluating only an exceedingly small fraction of the shapelets space.

%Our algorithm automatically outputs a classification model upon convergence to a high-accuracy model, thus making shapelets-based classification of large datasets feasible.

%Shapelets
%Problems
%Anytime
%Contributions 
\end{abstract}


\section{Introduction}
\label{sec:Intro}
A time series is a sequence of numerical data in which each item
is associated with a particular instance in time \cite{harris2003applied}.
These series are the focus of much research in diverse topics such as forecasting, database indexing, clustering and classification
\cite{agrawal1993efficient,faloutsos1994fast,liao2005clustering,geurts2001pattern}.
Time series classification is a field which has drawn much attention over the past decade
\cite{geurts2001pattern,kadous2005classification,keogh1998enhanced,povinelli2004time,ratanamahatana2004making,xi2006fast}.

A new approach for classification of time series, proposed by Ye and Keogh \cite{ye2011time}, uses shapelets.
A \emph{shapelet} is a subsequence extracted from one of the time series in the dataset.
The shapelet is chosen by its ability to split the data into two subsets,
such that as many time series as possible belonging to one class will be in one of the subsets.
Classifying a time series is done based on whether or not its distance from the shapelet
is below a pre-computed threshold value associated with the shapelet.
If more than a single shapelet is required, a classification tree is used, with a shapelet placed in each node of the tree.
The intuition behind this approach is that the information required to separate the classes is an intrinsic part of the time series behavior
expressed best by the measurement values themselves, instead of using summaries of the data.
The algorithm of Ye and Keogh considers all the subsequences of the dataset's time-series
in order to identify those shapelets that yield the optimal split.
Through the rest of this paper, we will refer to this algorithm as the \textit{YK algorithm}.

Two key advantages of classification with shapelets are its high accuracy
and the interpretability of the classification model learnt.
A disadvantage of this approach is the time required for building the classification tree.
The search for the best shapelet requires examining all subsequences of all lengths from all time series in the training set.
For a small dataset (e.g. 30 time series each with 300 measurements), it may require a few hours to learn the model,
as the number of potential shapelets which need to be checked is in the millions
(for the example given there are 1,336,530 subsequences which need to be checked).
For somewhat larger datasets (e.g. 50 time series each with 1000 measurements leading to 24,925,050 different subsequences) the time required can be measured in days or even weeks.

%Articles dealing with this problem
A number of approaches for reducing the time complexity of shapelet-based model generation have been introduced.
McGovern et al. \cite{mcgovernidentifying} propose an algorithm in which the data is first discretized,
and then all subsequences of a minimum predefined length are evaluated,
and those deemed best are concatenated to create better motifs for classification.
A different approach, proposed by Hartmann et al.,
uses an evolution strategy to reduce the number of subsequences tested \cite{hartmann2010prototype}.
Both these techniques introduce heuristics for reducing the number of subsequences tested.
A method allowing faster analysis of \emph{all} subsequences was introduced by Mueen et al. \cite{mueen2011logical}.
They presented two optimizations, one which can determine whether a subsequence is worth testing or not
and another which reduces the time required to test a subsequence which was not dropped in the first step.

%Our approach and its novelty.
\subsection{Our Contributions}

The merits of shapelet based classification motivated us to attempt reducing the time required for generating shaplet-based classification models, thus making use of this approach practical not only for the smallest of datasets.
We introduce the ShApLet SAmpling algorithm (henceforth called the SALSA algorithm) for fast computation of shapelet-based classification trees which does not examine all possible shapelets. Instead, the SALSA algorithm samples and evaluates shapelets according to a pre-determined evaluation order only as long as the quality of the new shapelets being examined keeps improving. As part of this process, the distance of each shapelet to all time series is calculated. To speed up the construction of the rest of the classification tree, these distances are saved to disk. Once the quality of new shapelets stops improving, SALSA ceases searching for better shapelets and proceeds to build the rest of the classification tree, using only those shapelets previously examined.

We tested three different shapelet evaluation orders on a number of datasets of varying size with the goal of determining which evaluation order of the shapelets space yields the fastest convergence to an accurate model.
The following evaluation orders were implemented and tested:
1) The \emph{simple} evaluation order iterates through all possible shapelet lengths, from the shortest to the longest, as is done by the YK algorithm.
2) The \emph{binary search} evaluation order starts with a quick sampling of the shapelets space,
by extracting only non-overlapping shapelets of each length.
It then iteratively evaluates shapelets of each length, while varying the overlap length of evaluated shapelets in a binary manner
(see Section \ref{sec:quasi} for more details).
3) The \emph{random} evaluation order picks a random permutation of the shapelets space and evaluates shapelets according to it, allowing shapelets of all lengths to be examined early on. We henceforth refer to the SALSA algorithm using the random evaluation order as the SALSA-R algorithm.

Our results clearly indicate that the random evaluation order yields the best results of all three approaches. The SALSA-R algorithm converges extremely quickly to a highly accurate model after evaluating only an exceedingly small fraction of the shapelets space. Moreover, it is noteworthy that the accuracy of the shapelets-based model does not monotonically grow as a function of the number of evaluated shapelets. Rather, it grows up to a point (depending on which evaluation order is used) and then starts to decline. Thus, the SALSA-R algorithm can return a model which is even more accurate than that returned by the YK algorithm, which uses all shapelets. We interpret this result as follows: building a model based on too many shapelets makes shapelets-based classification susceptible to overfitting \cite{Dietterich:1995:OUM:212094.212114}.

The time required by the SALSA-R algorithm to converge to a high-accuracy model is orders of magnitude smaller than that of the YK algorithm and, as our experimental evaluation demonstrates, the accuracy of the model it returns is always close to -- and often superior to -- that of the model returned by the YK algorithm.

Investigation into the reason why random sampling order works  so well reveals that there are multiple high-quality shapelets which can be used for building the classification tree, but that they are  not evenly distributed through out the shapelets space. From the datasets we analyzed, it seems that all high-quality shapelets are
concentrated in a tight cluster of shapelet lengths and all are extracted from the same area in the time series.
In light of these results, we believe that the SALSA-R algorithm may be useful in practice for fast generation of accurate shapelets-based classification trees, since an exhaustive search of the shapelets space is exceptionally time consuming.

It is important to note that we are not the first to introduce random shuffling into an algorithm searching for shapelets. The YK algorithm also creates a random permutation of all shapelets \emph{of the same length}. However, it evaluates shaplets \emph{in increasing order of their length}. SALSA-R, on the other hand, shuffles \emph{all shapelets of all lengths}. As our analysis in Section \ref{sec:results} establishes, this seemingly small difference is vital for finding high-quality shapelets early on, since high-quality shaplets are concentrated in a very tight range of shaplet-lengths. In addition, our algorithm automatically outputs a classification model upon convergence to a high-accuracy model whereas the YK algorithm examines all shapelets.

The rest of the article is arranged as follows:
In section \ref{sec:origAlgo} some definitions are presented as well as a brief description of the YK algorithm.
In section \ref{sec:quasi} we present our algorithm
and three different approaches for defining the order in which shapelets are evaluated.
Section \ref{sec:results} shows experimental results.
Last, section \ref{sec:conclusions} brings conclusions and presents questions which arise from this work. 
\section{Background}
\label{sec:origAlgo}
\subsection{Definitions}
Here a number of definitions essential for proper understanding of the article are presented.
The following definitions formally describe a time series and measures of similarity between time series.
\begin{mydef}
\label{def:TS}
	A time series $T$ of length $m$ is a series of $m$ successive measurements:
	\begin{displaymath}
		T=t_0,t_1,...,t_{m-1}
	\end{displaymath}
\end{mydef}

\begin{mydef}
\label{def:subseq}
	A subsequence $S$ of length $l$ extracted from time series $T$ at position $i$ is a time series of the following form:
	\begin{displaymath}
		%S=t_i,t_{i+1},...t_{i+l-1}
		S=T[i:i+l-1]
	\end{displaymath}
\end{mydef}

\begin{mydef}
\label{def:dist}
	The Euclidean distance between two time series $T,V$ of length $m$ is
	\begin{displaymath}
		dist_E(T,V)=\sqrt{\displaystyle\sum\limits_{i=0}^{m-1} (t_i-v_i)^2}
	\end{displaymath}
\end{mydef}

\begin{mydef}
\label{def:subTsDist}
	The distance between two time series $T,V$ of different lengths $|T|=m, |V|=m+j, j>0$
	is the minimal distance between $T$ and all $j$ subsequences of length $m$ in $V$
\end{mydef}

The ensuing definitions describe a method for measuring the order induced by a shapelet.
\begin{mydef}
\label{def:entropy}
	Let us assume a dataset $D$ with time series from $k$ different classes.
	We will denote the number of representatives from each class as $C_i$.
	The fraction of time series from each class is $p(C_i)$ and the entropy of the dataset is:
	\begin{displaymath}
		Ent(D)=-\displaystyle\sum\limits_{i=0}^{k-1} p(C_i)\log{p(C_i)}
	\end{displaymath}
\end{mydef}

\begin{mydef}
\label{def:IG}
	Given a dataset $D$ which is split into two subsets $D_1,D_2$
	such that $D_1 \cap D_2 = \emptyset$
	and $D_1 \cup D_2 = D$, the information gain (IG) is the difference between the entropy of $D$ and the weighted average entropy of $D_1,D_2$:
	\begin{multline*}
		IG(D,D_1,D_2)= \\ Ent(D)- (p(D_1)Ent(D_1)+p(D_2)Ent(D_2))
	\end{multline*}
\end{mydef}

It is worth noting that entropy measures order and that the smaller the entropy the more order there is.
This implies that if a certain split of the dataset leads to a more ordered system, the entropy of the subsets will drop
and the information gain will increase.

\subsection{The YK Shapelet Extraction Algorithm}
For completeness we briefly describe the YK algorithm as presented in \cite{ye2011time}.
First, we present the algorithm for two classes; we then extend the description to a multi-class data set.
Let $D$ be a dataset with two classes and $N$ time series.
The YK algorithm extracts all possible subsequences of every length
(from a minimal length, usually 3, to a maximal length which is usually the length of the shortest time series)
from every time series.
For each subsequence $S$, the distance to each time series is calculated, as defined in def.~\ref{def:subTsDist}.
Then, the time series are ordered by their distance from $S$.
Using this induced order, for every two adjacent time series, their average distance to $S$ is calculated.
We will refer to this average distance as the \textit{splitting distance}.
Each of the $N$ splitting distances defines two subsets, one containing all time series with a distance to $S$
smaller than or equal to the splitting distance, and the other containing all time series with a distance to $S$ greater than the splitting distance.
For every possible split into two subsets, the information gain is calculated.
If the current information gain is better than the best so far, the shapelet is kept along with the corresponding splitting distance.
Tie breaking is done by keeping the shapelet which gives a larger average distance between the two subsets
which is referred to as the \textit{margin}.
After checking all possible subsequences, the best shapelet and the corresponding splitting distance are returned.

This method can be easily extended to a multi-class problem by building a tree, with a shapelet and splitting distance in each node.
A new node receives one of the two subsets created by the shapelet found by the node above it,
and learns the best shapelet and splitting distance for this subset of time series.
The stopping criteria for this recursive algorithm is when all the time series in the subset are from one class.

Two important implementation issues are that all distance calculations are done after local normalization,
and that the margin is normalized, by dividing it by the length of the subsequence.

Classification of a time series $t$ is accomplished by walking through the tree.
At each node the distance of the shapelet $S$ associated with the node to $t$ is calculated.
The node decides to which of its children $t$ should be directed,
depending on whether its distance from $S$ is smaller or greater than the splitting distance.
When $t$ reaches a leaf, it is assigned the class associated with this leaf. 
\section{The Shaplet Sampling (SALSA) Algorithm}
\label{sec:quasi}
\subsection{Algorithm Description}
\label{subsec:algorithm}
%An intuitive overview of the algorithm
The goal of our approach is to provide an accurate classification model as soon as possible.
To achieve this, we introduced four major changes to the YK algorithm.
First, we changed the order in which shapelets are examined, to allow fast sampling of the entire shapelet domain.
Second, our algorithm samples and examines only a small subset of all possible shapelets,
ceasing analysis of new shapelets once the quality of shapelets stabilizes.
We measure stability by inspecting whether new shapelets improve the IG and margin of the current best shapelet
by a significant amount or not, indicating whether a shapelet similar to that with the best IG and margin has been found.
Third, while examining shapelets in the root node, the distances of each shapelet to all time series are saved to disk.
Last, while building the rest of the classification tree, only those shapelets already examined by the root,
with distances to time series already precalculated, are considered as candidates.
This eliminates the need to recalculate distances of shapelets to time series in each node.

%A detailed explanation of the algorithm
We describe the SALSA algorithm as a two stage procedure.
The first stage, described in procedure~\ref{proc:rootExtractShapelets}, searches for the best shapelet for the root node.
The second stage, outlined in procedure~\ref{proc:nodeExtractShapelets}, builds the rest of the tree.
The root requires a separate algorithm as the distances of subsequences to time series are as yet unknown,
while in the rest of the nodes, these distances have already been calculated.
In addition, the root must be able to identify the stabilization of shapelet quality,
while the rest of the nodes check all subsequences already considered by the root.

\begin{algorithm}[t]
\floatname{algorithm}{Procedure}
\begin{algorithmic}[1]
\item[RootShapeletSearch($D,\epsilon,NI$)]
\COMMENT {Input is a dataset and stabilization parameters}
\item[]
\STATE $E \gets enumerator\ of\ shapelets$ \label{algoLine:queue}
\STATE Initialize $tree$
\STATE $sDst \gets \emptyset$ \COMMENT {Distances of shapelets from time series}
\STATE $iCnt \gets 0$ \COMMENT{Number of iterations shapelet quality didn't improve} \label{algoLine:endInit}
\item[]
\REPEAT \label{algoLine:beginMain}
	\STATE $s \gets E.next()$ \COMMENT {next shapelet to examine}
	\FOR {time series $t$ in $D$}
		\STATE $d \gets $ $dist_E(s,t)$
		\STATE $sDst$.push($<s,t,d>$)
	\ENDFOR
	\STATE Calculate quality of $s$
	\IF {shapelet $s$ better than best}
			\STATE Save (s, splitting distance, IG, margin)
	\ENDIF
	\IF {shapelet quality better than best by more than $\epsilon$}
		\STATE $iCnt \gets 0$
	\ELSE
		\STATE $iCnt++$
	\ENDIF
	
\UNTIL{$iCnt==NI$ or all shapelets have been checked}\label{algoLine:endMain}
\item[]
\STATE $leftD,rightD \gets splitDataset$ \label{algoLine:beginBuild}
\STATE $tree.leftTree \gets$ BuildSubTree($leftD,sDst$)
\STATE $tree.rightTree \gets$ BuildSubTree($rightD,sDst$) \label{algoLine:endBuild}
\RETURN $tree$
\end{algorithmic}
\caption{Search for shapelet in Root node}
\label{proc:rootExtractShapelets}
\end{algorithm}

In procedure~\ref{proc:rootExtractShapelets}, the input is a dataset of time series $D$ and two parameters defining stabilization:
$\epsilon$ and \emph{NI}. \emph{NI} (Number of Iterations) defines the number of shapelet-evaluation iterations after which the search for a better shapelet in the root node should terminate in the lack of significant quality improvement. If a shapelet substantially better than the previous best is found, the count is reset to $0$.
The amount of change in shapelet quality regarded as substantial is specified by $\epsilon$. Shaplet sampling proceeds if the IG increases by a ratio of at least $\epsilon$ or if the IG remains the same but the margin increases by a ratio of at least $\epsilon$ in the course of the last \emph{NI} iterations.
During initialization (lines \ref{algoLine:queue}-\ref{algoLine:endInit}) an enumerator $E$ is initialized, that specifies the sampling order of shapelets; that is, the enumerator defines the order in which the time series subsequences are examined during the algorithm's execution.
The sampling order determined by $E$ is the subject of section~\ref{subsec:OrderOfSubSeq}. Also in this phase, a data structure to contain the distances of all time series to all shapelets, $sDst$, is initialized,
as is a counter $iCnt$, counting the number of iterations since the last substantial change in shapelet quality.

The next segment of code (lines \ref{algoLine:beginMain}-\ref{algoLine:endMain}) describes the analysis of a shapelet
by calculating its distance from all time series,
determining its ability to split the dataset meaningfully (IG and margin)
and evaluating whether it should substitute the current shapelet.
In addition, book keeping is done, to promise that distances of subsequences to time series already calculated can be reused by other nodes.

Last, after either stabilization of shapelet quality or examination of all shapelets,
the rest of the tree is built (lines \ref{algoLine:beginBuild}-\ref{algoLine:endBuild}).
The dataset is split into two subsets using the splitting distance.
All time series with a distance to the chosen shapelet smaller than the learnt splitting distance are assigned to the left subset
and the rest are allocated to the right one.
Then, for each subset a subtree is built using procedure~\ref{proc:nodeExtractShapelets}.

Procedure~\ref{proc:nodeExtractShapelets} requires a dataset $D$
as well as the data structure containing all shapelets examined and their distances to all time series.
As the procedure is recursive (lines \ref{algoLine:recursiveLeft}-\ref{algoLine:recursiveRight}), the base case is checked first,
returning a leaf if all time series in $D$ are from one class (lines \ref{algoLine:baseBegin}-\ref{algoLine:baseEnd}).

Next, all shapelets examined by the root, which were extracted from $D$
(as $D$ is only a subset of the original dataset,
some of the shapelets may have been extracted from time series not in this particular subset)
are examined in the same fashion as in procedure~\ref{proc:rootExtractShapelets}.

Two special edge cases are if none of the shapelets in $sDst$ were extracted from time series in subset $D$
(lines \ref{algoLine:edge1Begin}-\ref{algoLine:edge1End})
or if the best shapelet found by this node couldn't split $D$ into two non-empty subsets
(lines \ref{algoLine:edge2Begin}-\ref{algoLine:edge2End}).
In both these cases, a leaf is created as an attempt to build a sub-tree would lead to infinite recursion.
The class assigned is decided by a majority vote (i.e. the class with the most representatives)
breaking ties by choosing the class with the smaller value.

\begin{algorithm}[t]
\floatname{algorithm}{Procedure}
\begin{algorithmic}[1]
\item[BuildSubTree($D$,$sDst$)]
\COMMENT {Input is a dataset and all shapelets already examined with their distances all time series}
\item[]

\STATE Initialize $tree$
\IF {all time series in $D$ are from one class}\label{algoLine:baseBegin}
	\RETURN Leaf representing class\label{algoLine:baseEnd}
\ENDIF

\item[]
\REPEAT
	\STATE $s \gets sDst.next()$ \COMMENT {next shapelet to examine}
	\IF {$s$ was extracted from $D$}
		\STATE Calculate quality of $s$
		\IF {shapelet $s$ better than best}
			\STATE Save (s, splitting distance, IG, margin)
		\ENDIF
	\ELSE
		\STATE continue
	\ENDIF
\UNTIL{all shapelets have been checked}

\item[]
\IF {$sDst$ contains no shapelets extracted from $D$}\label{algoLine:edge1Begin}
	\STATE Decide class by majority vote
	\RETURN Leaf representing class\label{algoLine:edge1End}
\ENDIF
\IF {$D$ couldn't be split by best shapelet}\label{algoLine:edge2Begin}
	\STATE Decide class by majority vote
	\RETURN Leaf representing class\label{algoLine:edge2End}
\ENDIF
\item[]
\STATE $lDataset,rDataset \gets splitDataset$
\STATE $tree.leftTree \gets$ BuildSubTree($lDataset,sDst$)\label{algoLine:recursiveLeft}
\STATE $tree.rightTree \gets$ BuildSubTree($rDataset,sDst$)\label{algoLine:recursiveRight}
\RETURN $tree$
\end{algorithmic}
\caption{Building a sub-tree}
\label{proc:nodeExtractShapelets}
\end{algorithm}

\subsection{Shapelets Sampling Order}
\label{subsec:OrderOfSubSeq}
Here we elaborate on line~\ref{algoLine:queue} of procedure~\ref{proc:rootExtractShapelets},
which introduced an enumerator of the shapelets, determining the order in which shapelets are considered by the algorithm.
Following are three different procedures, each specifying a different sampling order. The pseudo-code of these procedures specifies the order in which the shapelets domain is enumerated. In other words, calls of the \emph{next}
method implemented by an enumerator output shapelets in the order prescribed by the corresponding
procedure.

The first enumeration order, sketched in procedure~\ref{proc:simple}, simply iterates through all possible shapelet lengths from shortest to longest,
extracting all shapelets from each time series.
The second, described in procedure~\ref{proc:fast}, samples the entire shapelets domain quickly,
by extracting only non-overlapping shapelets of each length in every iteration of the outer loop (line \ref{algoline:fastrepeat})
instead of extracting all shapelets of each length at once.
An important parameter is \emph{startIndex},
which defines the index in the time series from which we start extracting the non-overlapping subsequences
(each start index between 0 and the length of the subsequence will define a different set of non-overlapping subsequences).
We defined the function nextStartIndex (line~\ref{algoline:fastNextIndex})
to enhance the procedure's ability to promise fast coverage by choosing $startIndex$ intelligently.
Iteration over the start indices is done in a manner resembling a binary search, hence the procedure's name,
always defining the next start index as the middle between two previously examined start indices.
For example, if extracting subsequences $8$ measurements long, the order of the start indices would be as follows:
$[0,4,2,6,1,3,5,7]$. The first start index is $0$. The next start index is calculated as halfway between examined indices (0,8) etc.
This combination of extraction of non-overlapping subsequences and the special way in which the next start index is determined are meant to ensure that we quickly sample the entire subsequence domain and that a subsequence very similar to the best shapelet will be encountered early on.
The last method, depicted in procedure~\ref{proc:random},
enumerates shapelets according to the order defined by a random permutation of the entire shapelets space.

%The minimum length of a shapelet was defined as $3$, as we locally normalize all subsequences and this is the shortest %length which has meaningful information.
%(Locally normalizing subsequences of length 2 leaves us with
%either $[1,-1]$ or $[-1,1]$.) The maximum length is simply the length of the shortest time series in the dataset.

\begin{algorithm}[t]
\floatname{algorithm}{Procedure}
\begin{algorithmic}[1]
\item[SimpleSearch($D$)]
\COMMENT {Input is a dataset}
\item[]
\STATE $minLen \gets 3$
\STATE $maxLen \gets $ length of shortest time series
\item[]
\FOR {subsequence length $l$ from $minLen$ to $maxLen$}
	\FOR {time series $t$ in $D$}
		\STATE $lastIndex \gets$ (\emph{t.length - l +1})
		\FOR {index $i$ from $0$ to $lastIndex$}
			\STATE $subseq \gets t[i:i+l-1]$
			\STATE output($subseq$)
		\ENDFOR
	\ENDFOR
\ENDFOR
\RETURN $E$
\end{algorithmic}
\caption{Simple enumeration of shapelets domain}
\label{proc:simple}
\end{algorithm}

\begin{algorithm}[h]
\floatname{algorithm}{Procedure}
\begin{algorithmic}[1]
\item[binary($D$)]
\COMMENT {Input is a dataset}
\item[]
\STATE $minLen \gets 3$
\STATE $maxLen \gets $ length of shortest time series

\item[]
\REPEAT \label{algoline:fastrepeat}
\FOR {subsequence length $l$ from $minLen$ to $maxLen$}
	\IF{all subsequences of this length have been extracted}
		\STATE continue
	\ENDIF
	\STATE $startIndex \gets$ nextStartIndex()\label{algoline:fastNextIndex}
	\FOR {time series $t$ in $D$}
		\STATE $lastIndex \gets$ (\emph{t.length - l +1})
		\FOR {index $i =0;\ i \leq lastIndex;\ i=i+l$}
			\STATE $subseq \gets t[i:i+l-1]$
			\STATE output($subseq$)
		\ENDFOR
	\ENDFOR
\ENDFOR
\UNTIL{all subsequences have been extracted}
\RETURN $E$
\end{algorithmic}
\caption{Fast-coverage enumeration of shapelets domain}
\label{proc:fast}
\end{algorithm}

\begin{algorithm}[h]
\floatname{algorithm}{Procedure}
\begin{algorithmic}[1]
\item[RandomSearch($D$)]
\COMMENT {Input is a dataset}
\item[]
\STATE $P \gets$ random permutation of shapelets domain
\STATE output shaplets according to the order defined by $P$
\end{algorithmic}
\caption{Random enumeration of shapelets domain}
\label{proc:random}
\end{algorithm}

%Describe memory usage of the different methods (random with repitions doesn't require much memory.) - I decided not to address this, as medium sized datasets it shouldn't be a problem in languages like C. For large datasets it can probably be handled by clever writing to disk.

%Describe the existing optimizations which can be used (First from the first of Keogh's articles (pruning may be problematic). From the second article speed up can be used. pruning can be used only with simple).
In \cite{ye2011time,mueen2011logical}, optimizations for the original algorithm were introduced.
Some accelerate the distance calculations and some allow pruning of shapelets before their distance to all time series
have been computed.
The SALSA algorithm can encompass any of the optimizations accelerating the distance calculations,
but cannot include those which prune shaplets
as pruning  methods can reject a shapelet before its distance to all time series has been calculated.
In the SALSA algorithm, the distances of the shapelets to the time series are required not only by the root
but also by the other nodes in the tree: a shapelet that is useless in the root may be the best shapelet for a subset of the dataset assigned to one of the inner nodes.
%Therefore, our method requires that the distance of subsequences to all time series must be calculated preventing the %use of pruning methods.

\section{Experimental Results}
\label{sec:results}
%Describe the purpose of the experiments. (fast convergence, stability, reaches Keogh's result in the end, best way to converge quickly)
The purpose of our experiments was threefold.
First, we compared the performance of the three different enumeration orders in which shapelets are examined, described in section~\ref{subsec:OrderOfSubSeq}.
Our objective was to find which enumeration order allows returning the most accurate model fastest.
%By accurate we mean (here and in the rest of the article)
%in comparison with the accuracy achieved by the YK algorithm and not in absolute terms.
Second, after having found that the randomized evaluation order is significantly better than the alternative evaluation orders, we analyzed the distribution of high-quality shapelets within the datasets at hand in order to gain insights into the reasons underlying that. Third, we analyzed the performance and accuracy of the SALSA algorithm using the random evaluation order (SALSA-R) and compared it with those of the YK algorithm.

%Describe the setup - platform for running experiments, datasets.
\subsection{Comparison of Shaplets Evaluation  Orders}
\subsubsection{Experimental Setup}
\label{subsubsec:setup}
We tested the three different shaplet evaluation orders on all one-dimensional datasets on which the YK algorithm was assessed: arrowhead, coffee, mallat, shield and wheat. Two additional datasets we used were ecgPatterns and controlCharts. We split ecgPatterns into two, using the first half of the dataset for training and the second half for testing. All datasets are available online
\cite{ecgDataset},\cite{shapeletDatasets},\cite{ucrDataset}\footnote{controlCharts appears as synthetic controls}.
Table~\ref{table:datasets} summarizes important features of the datasets we used:
the size of a dataset refers to the number of time series
and the time series length measures the length of the shortest time series in the training set.
As can be seen, the training sets vary in size and in number of classes.

The arrowhead dataset describes different types of arrowheads classified by the geographic location in which they were found. The shield dataset contains shapes of shields used in different places and during different eras, as depicted in historical documents. Wheat and coffee both contain spectrographs of different strains, with a goal of distinguishing between them. ControlCharts and mallat are both synthetically generated datasets, used for testing different classification methods.

\begin{table}
\centering
\caption{Information on Datasets}
\label{table:datasets}
\begin{tabular}{|c|c|c|c|c|} \hline
Dataset		&train set 			&test set 			& \#classes	& time series 			\\
					&	size					&size						&						& length						 \\	\hline
arrowhead &36							&	175						&	3					&	340								 \\ \hline
coffee		&28							&	28						& 2					&	286								 \\ \hline
controlCharts &	200				&	400						& 6					& 60								\\ \hline
ecgPatterns	&	100					&	100						&	4					&	40								 \\ \hline
mallat 		&	320						&	2080					& 8					&	256								 \\ \hline
shield		&	30						&	129						& 3					&	994								 \\ \hline
wheat			&	49						&	726						&	7					&	1050							 \\ \hline
\end{tabular}
\end{table}

To assess how the accuracy of the model evolves over time
using each of the three methods for ordering the shapelets,
we implemented the YK algorithm
%(instead of using our stabilization algorithm)
and tested its performance as it evaluates shapelets in the order prescribed by these methods. We halt the algorithm at a pre-determined number of \emph{test points}. At each such test point, the process of examining new shapelets in the root is halted and a classification tree using only the shapelets already examined is built. Test points are set so that the number of shaplets evaluated between every two test points is equal.
%Parameter \emph{numSh} is determined at the beginning of the program's execution in the following manner.
%First, the total number of shapelets in a dataset is computed analytically. Then, \emph{numSh} is calculated by %dividing the total number of shapelets by the number of interrupts required.
The accuracy of the models built at these test points is determined based on the dataset's test set.

For mallat, we use 50 test points and for the rest of the datasets we use about 500 test points.
The reason for evaluating performance on mallat using only 50 test points is that due to its size, significantly more time is required to build the classification tree as compared with the other data sets. Results for the random evaluation order can vary, therefore we executed the experiment 10 times and averaged the accuracy at each test point.

%Describe the measures used
\subsubsection{Performance Measures}
\label{subsubsec:perfomance}
Evaluation of the classification trees was done by computing the percentage of time series which were correctly classified
from all time series in the test set.

Comparing performance of different algorithms requires special consideration.
In many cases, simply comparing algorithms' running times does not seem to be the best measure as it highly depends on the runtime environment used,
and on the quality of the code used to implement the algorithm. Consequently, many works use a measure that abstracts away these factors \cite{chen2004marriage,ding2008querying,wei2005atomic}.
We take a similar approach. Our experimental evaluation uses the number of shapelets examined by an algorithm as a measure of the work done by it. We thus assess the accuracy obtained by an algorithm, as a function of the number of shapelets it evaluates.

\subsubsection{Examination of Ordering Methods}
\label{subsubsec:results}
\begin{figure*}
	\centering
	\subfigure[arrowhead]{\label{fig:arrowhead}\epsfig{file=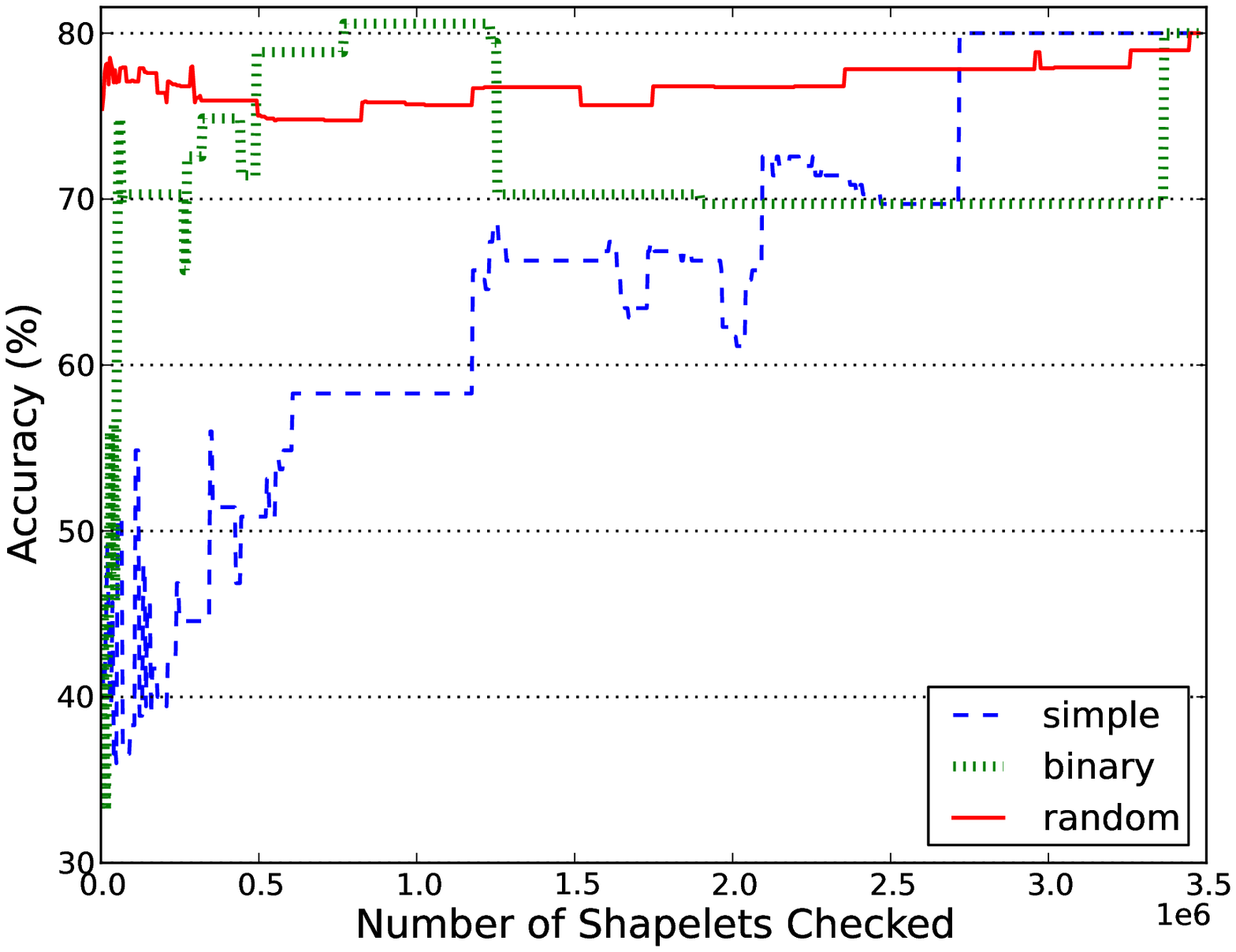,width=55mm}}
	\subfigure[coffee]{\label{fig:coffee}\epsfig{file=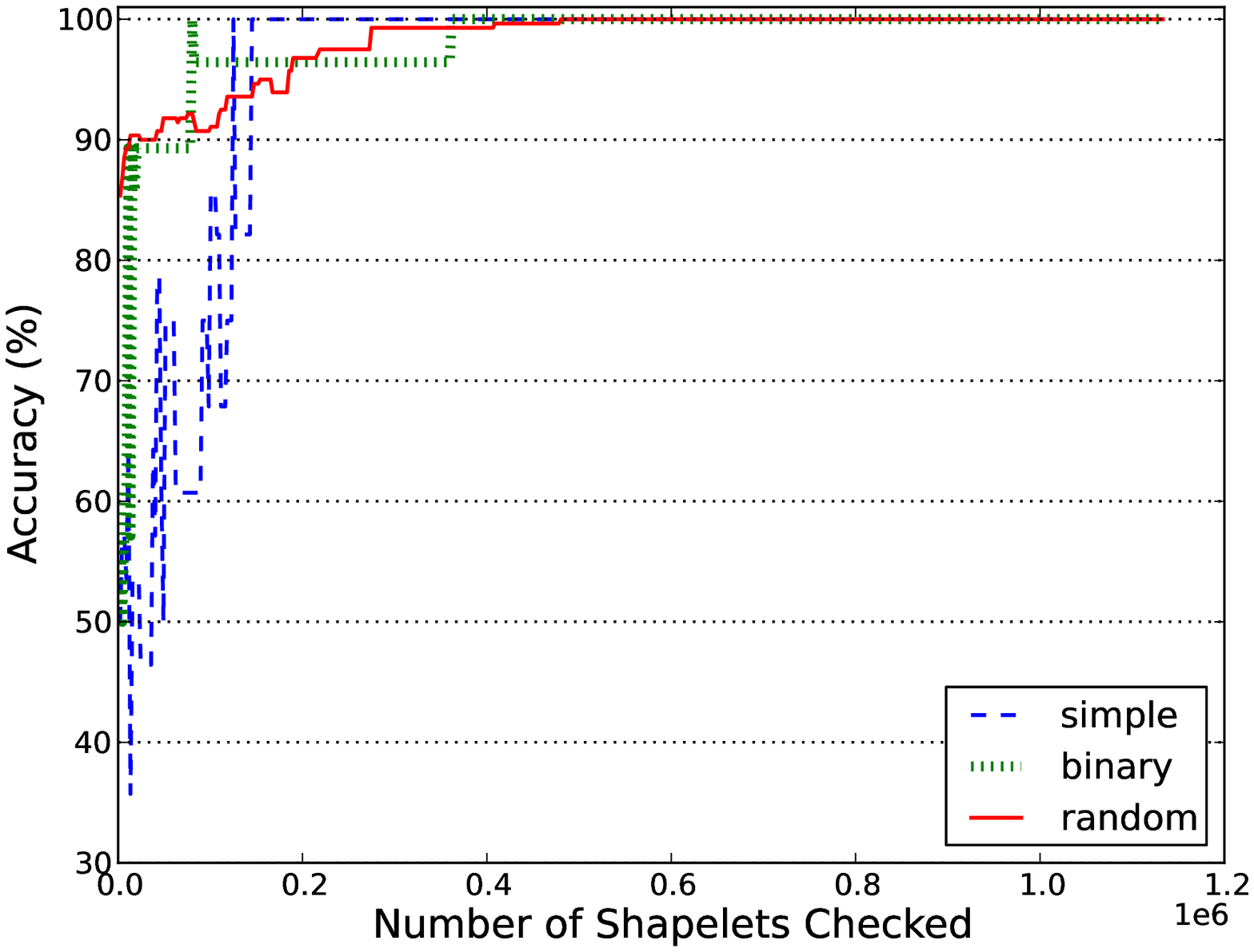,width=55mm}}
	\subfigure[controlCharts]{\label{fig:ControlCharts}\epsfig{file=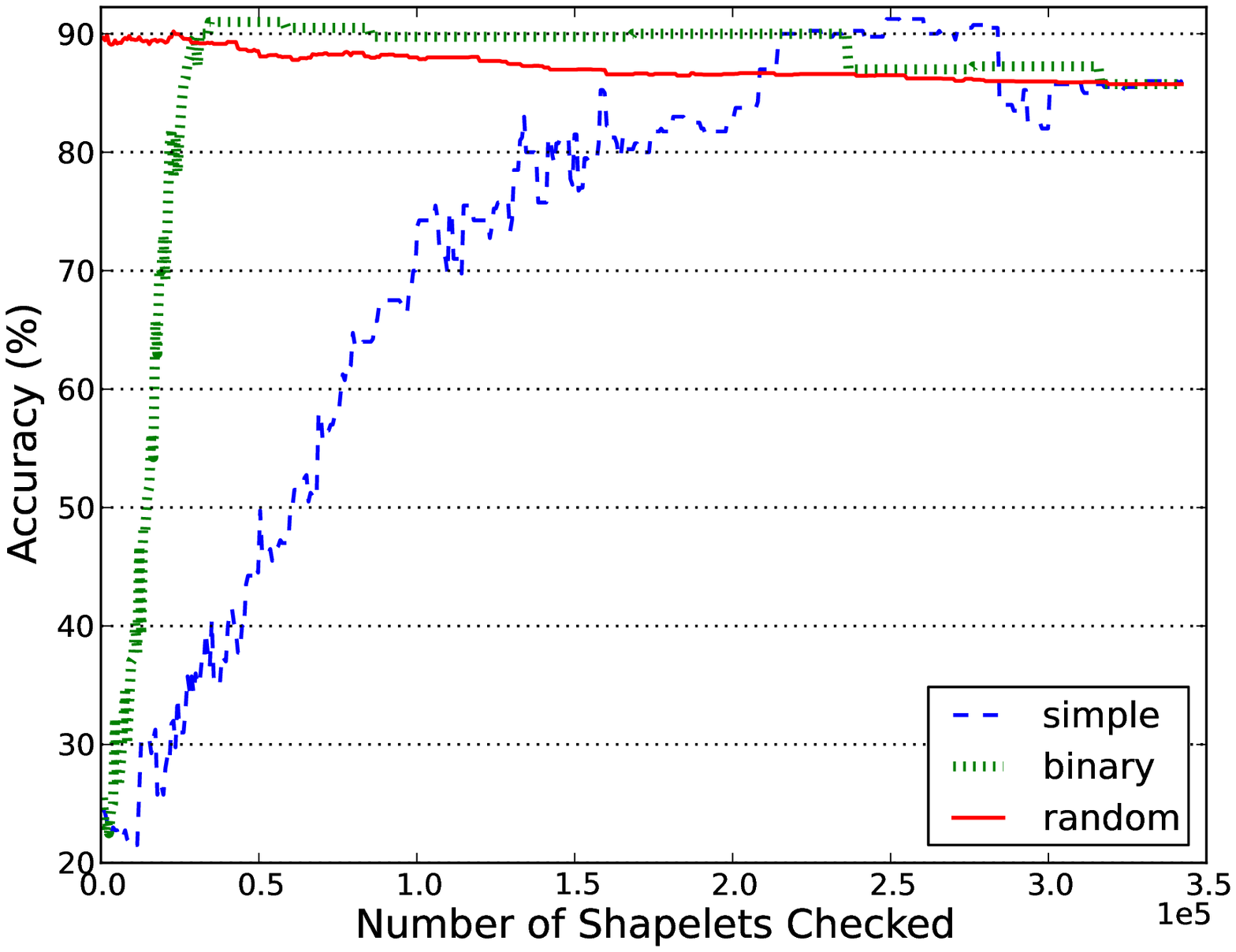,width=55mm}}
	\subfigure[ecgPatterns]{\label{fig:ecgPatterns}\epsfig{file=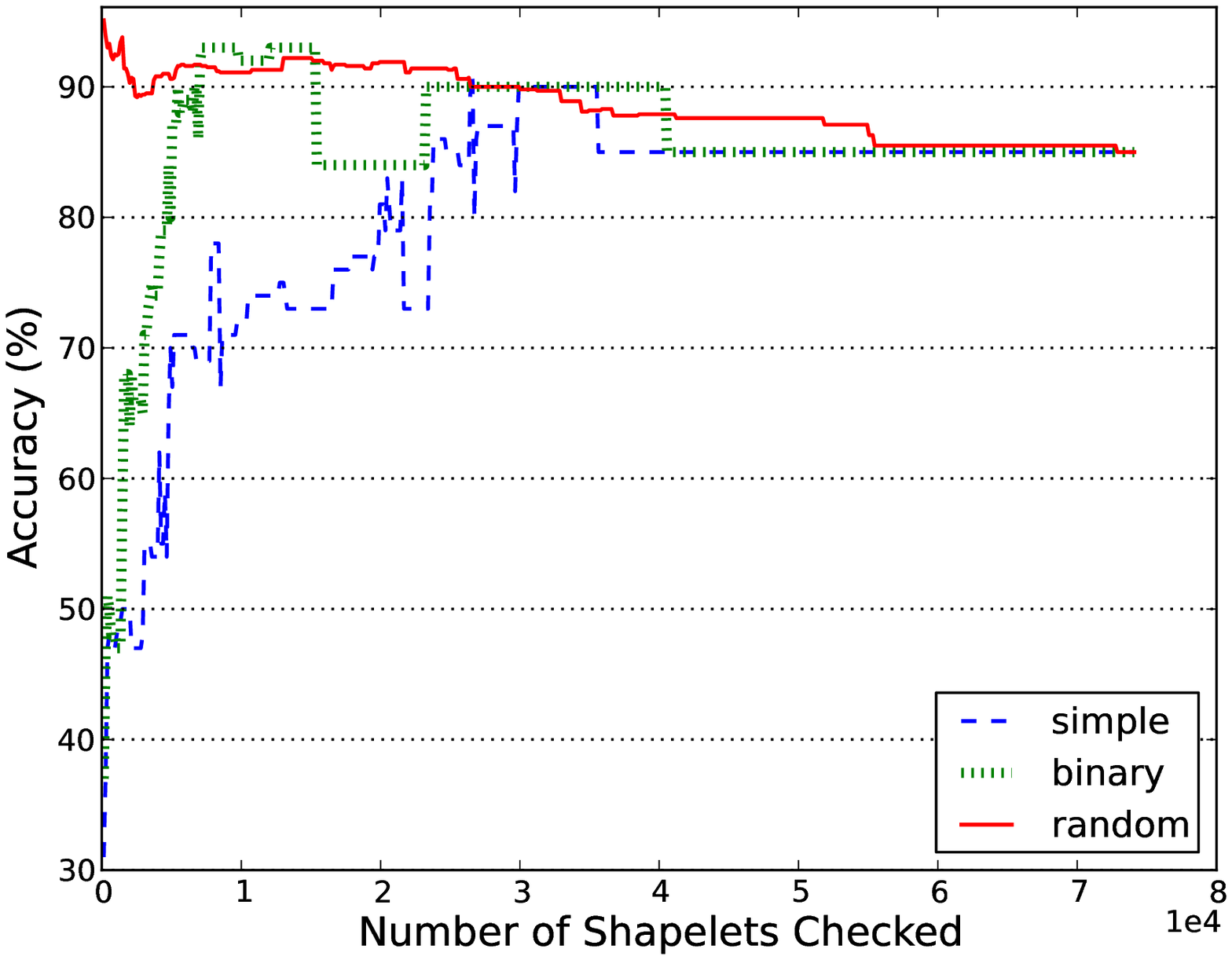,width=55mm}}
	\subfigure[mallat]{\label{fig:mallat}\epsfig{file=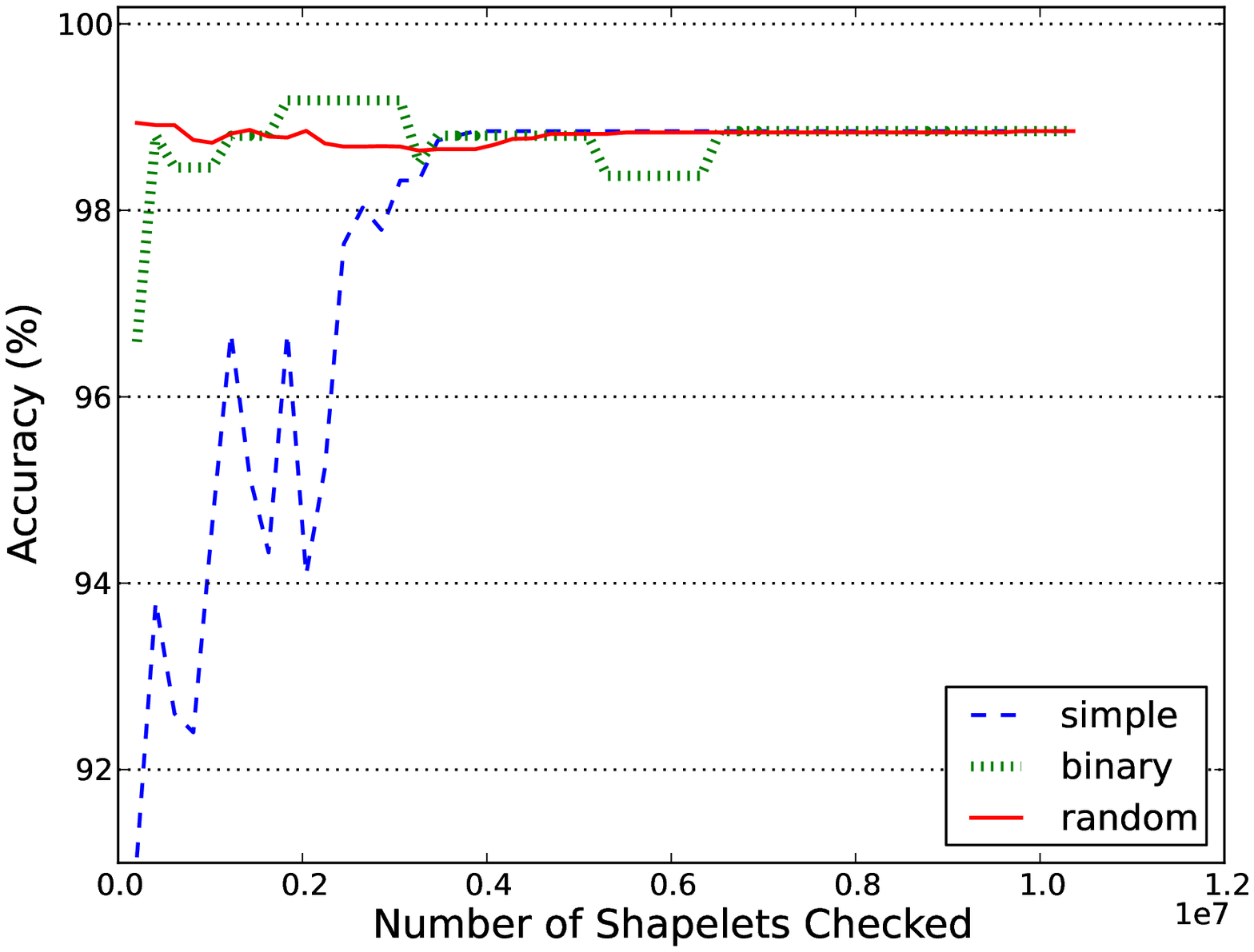,width=55mm}}
	\subfigure[shield]{\label{fig:shield}\epsfig{file=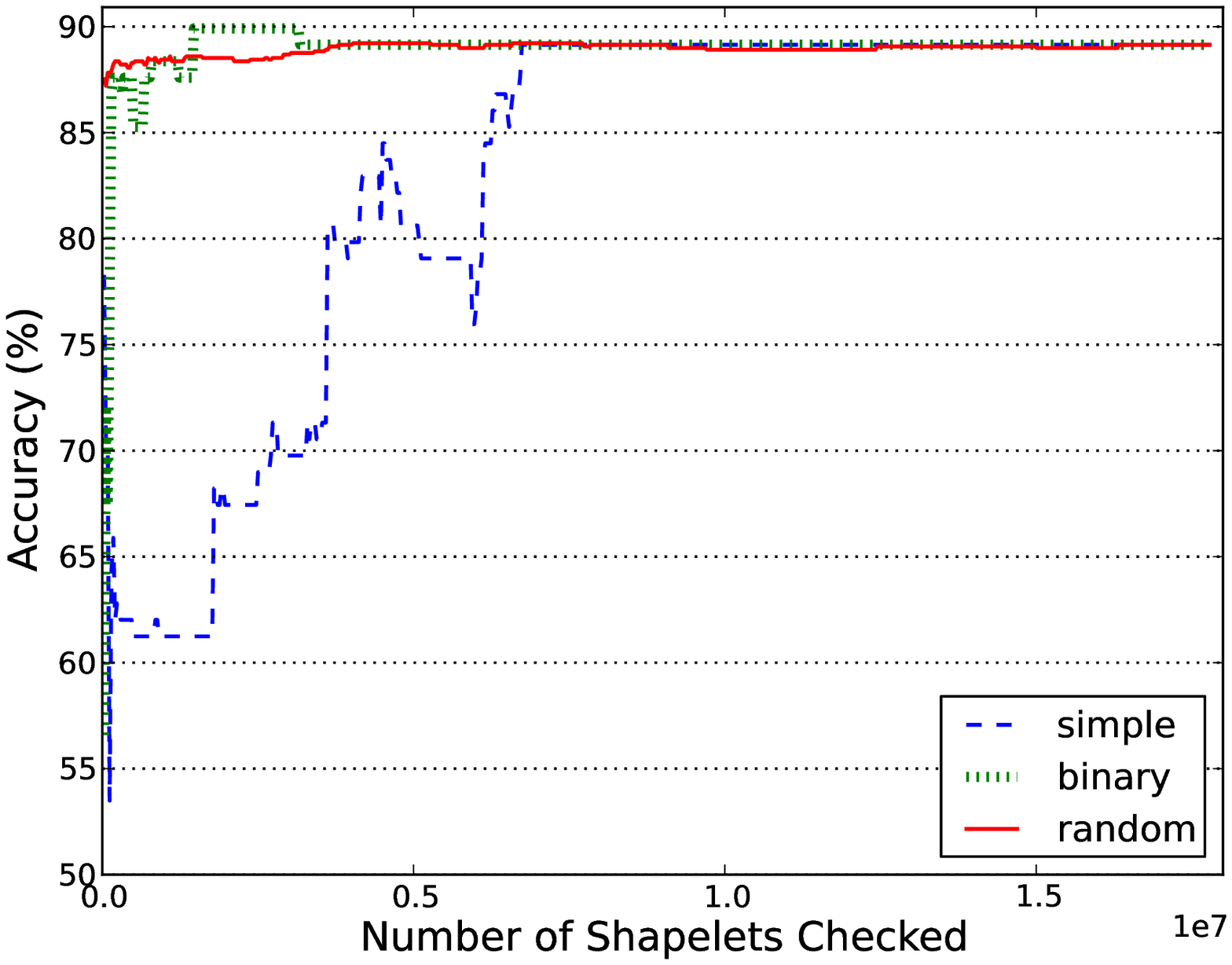,width=55mm}}
	\subfigure[wheat]{\label{fig:wheat}\epsfig{file=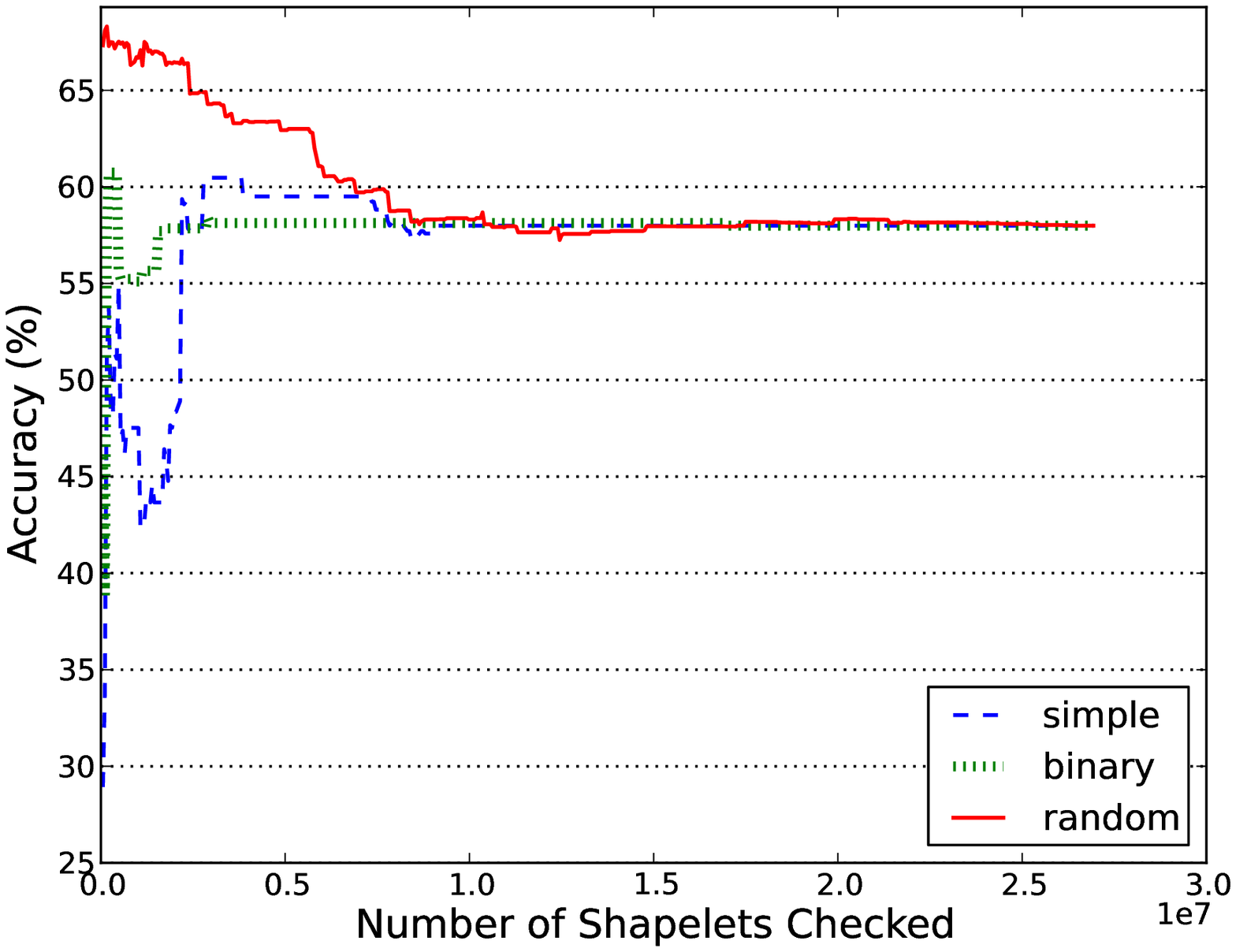,width=55mm}}
	\caption{Comparison of three enumeration methods for searching for the best shapelet.
	Blue is \textit{simple}, green is \textit{binary} and red is \textit{random}.}
	\label{fig:comparison}
\end{figure*}

%Our goal was to return a good model quickly.
%To achieve this goal, we tested three different approaches for the order in which the shapelet domain should be %searched.
%For each dataset and ordering method, we interrupted the search for the best shapelet 500 times
%(for mallat dataset 50 times) and measured the accuracy.\\
Fig.~\ref{fig:comparison} shows the results of our experiments.
For each dataset, there is a separate diagram showing the accuracy of each of the orderings as a function of the number of shapelets examined.
We ran the \emph{random} method 10 times, the red line showing the average accuracy at each point.
The graphs clearly show that, whereas the initial accuracy of \textit{simple} and \textit{binary} are rather low,
the initial accuracy of \textit{random} is extremely close to the accuracy after searching the entire shapelets space\footnote{The accuracy of the final model is identical to that obtained by the implementation of the YK algorithm.},
in many cases even surpassing it.

The graphs also show that accuracy does not necessarily increase with the number of shapelets examined,
although the trend is usually positive.
Declines in accuracy after inspecting additional shapelets can be seen in all datasets.
Moreover, for \textit{random} the trend is in some cases negative,
with the accuracy of the first models outperforming that of the final model
(Fig.~\ref{fig:comparison}~\subref{fig:ControlCharts},\subref{fig:ecgPatterns},\subref{fig:wheat}).
These declines in accuracy can be interpreted as a manifestation of overfitting \cite{Dietterich:1995:OUM:212094.212114}
which states that searching for the model which best fits the training set may memorize peculiarities of the training set,
instead of finding a more general rule.
%For our problem, the number of dimensions is the number of shapelets examined.
%This phenomenon states that as the number of dimensions increases,
%the size of the training set required to select the best features also increases.
%Therefore, as the number of shapelets examined grows, the ability of our algorithm to distinguish between shapelets decreases,
%leading to situations in which a shapelet is substituted by a less informative one.\\

Averaging many samples smoothes the curvature of a graph.
As we averaged the results from 10 executions of the \textit{random} method,
the smooth behavior of the graph showing the accuracy of the \emph{random} method may be a
side effect of averaging and not represent the true behavior of the algorithm.
We performed a simple test presented in Table~\ref{table:minMax},
to check whether there are sharp fluctuations in accuracy obtained by our \emph{random} method, not seen in the graphs due to smoothing.
For each dataset, from all 10 experiments, we chose the minimal and maximal accuracy obtained \emph{anywhere during the experiment}.
For the majority of datasets the minimal accuracy is within 10\% of the final accuracy
(that obtained after all shapelets have been examined) and the maximal accuracy usually surpasses it.
It is therefore safe to say that our averaging does not hide sharp fluctuations in accuracy.

Clearly from Fig.~\ref{fig:comparison}, random shapelet sampling converges extremely fast to an accurate model and invariably provides models that are significantly superior to those obtained by the other evaluation orders for the first test points.

%Mention that all results were compared with YK algorithm as they sent us.

\begin{table}
\centering
\caption{Variance in Random}
\label{table:minMax}
\begin{tabular}{|c|c|c|c|} \hline
Dataset				&Min		&Max		&Final	\\ 	\hline
arrowhead			&64.0\%	&82.9\%	&80.0\%	\\ 	\hline
coffee				&78.6\%	&100.0\%&100.0\%\\ 	\hline
controlCharts &81.7\%	&93.5\%	&85.7\%	\\ 	\hline
ecgPatterns		&83.0\%	&98.0\%	&85.0\%	\\ 	\hline
mallat				&98.0\%	&99.6\%	&98.8\%	\\	\hline
shield				&84.5\%	&90.7\%	&89.1\%	\\ 	\hline
wheat					&52.3\%	&72.6\%	&58.0\%	\\	\hline
\end{tabular}
\end{table}

\subsection{Analysis of High-Quality Shapelets Distribution}
\label{subsec:randSuc}
%As shown in \ref{sec:results} random ordering of the shapelets allows building of accurate models
%without examining large portions of the shapelet domain.
To understand why random sampling converges so quickly to a high-accuracy model, we measured the quality of all shapelets (in terms of information gain (IG) and margin) in each of the datasets
described in table \ref{table:datasets}. We extracted all shapelets with IG within $10\%$ of and margin within
$40\%$ of those of the best shapelet for the root. We chose these bounds on shapelet quality,
as we noticed that in the previous experiment all models with high accuracy came from within this range.

The focus of our analysis was to determine whether high-quality shapelets are evenly distributed throughout the shapelet space, or whether they are concentrated in certain areas. In this context, even distribution means that shapelets of all lengths, from any position in the time series, have equal probability of achieving good quality measures.

\begin{figure*}
	\centering
	\subfigure[arrowhead]{\label{fig:arrowhead_hist}\epsfig{file=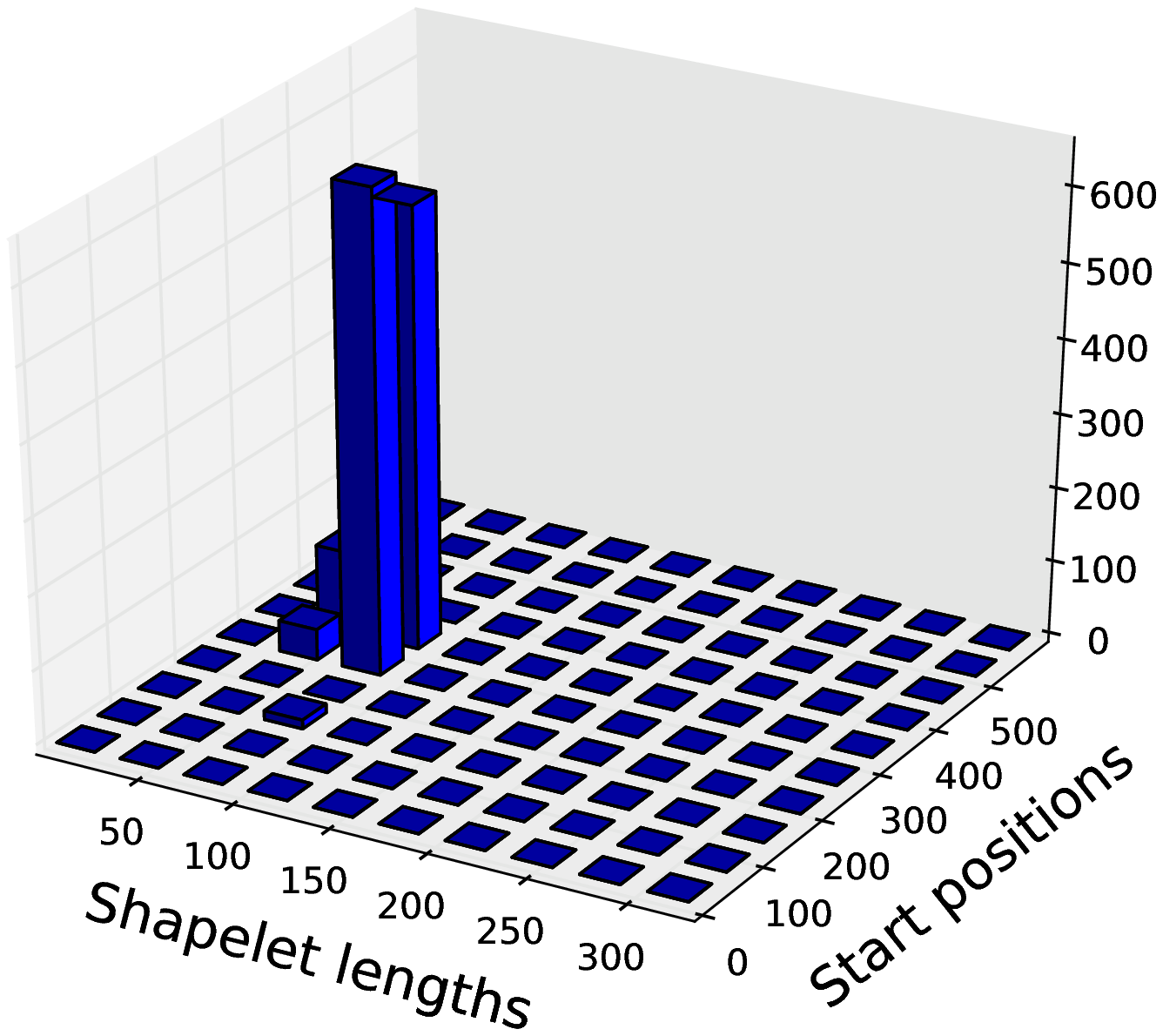,width=53mm}}
	\subfigure[coffee]{\label{fig:coffee_hist}\epsfig{file=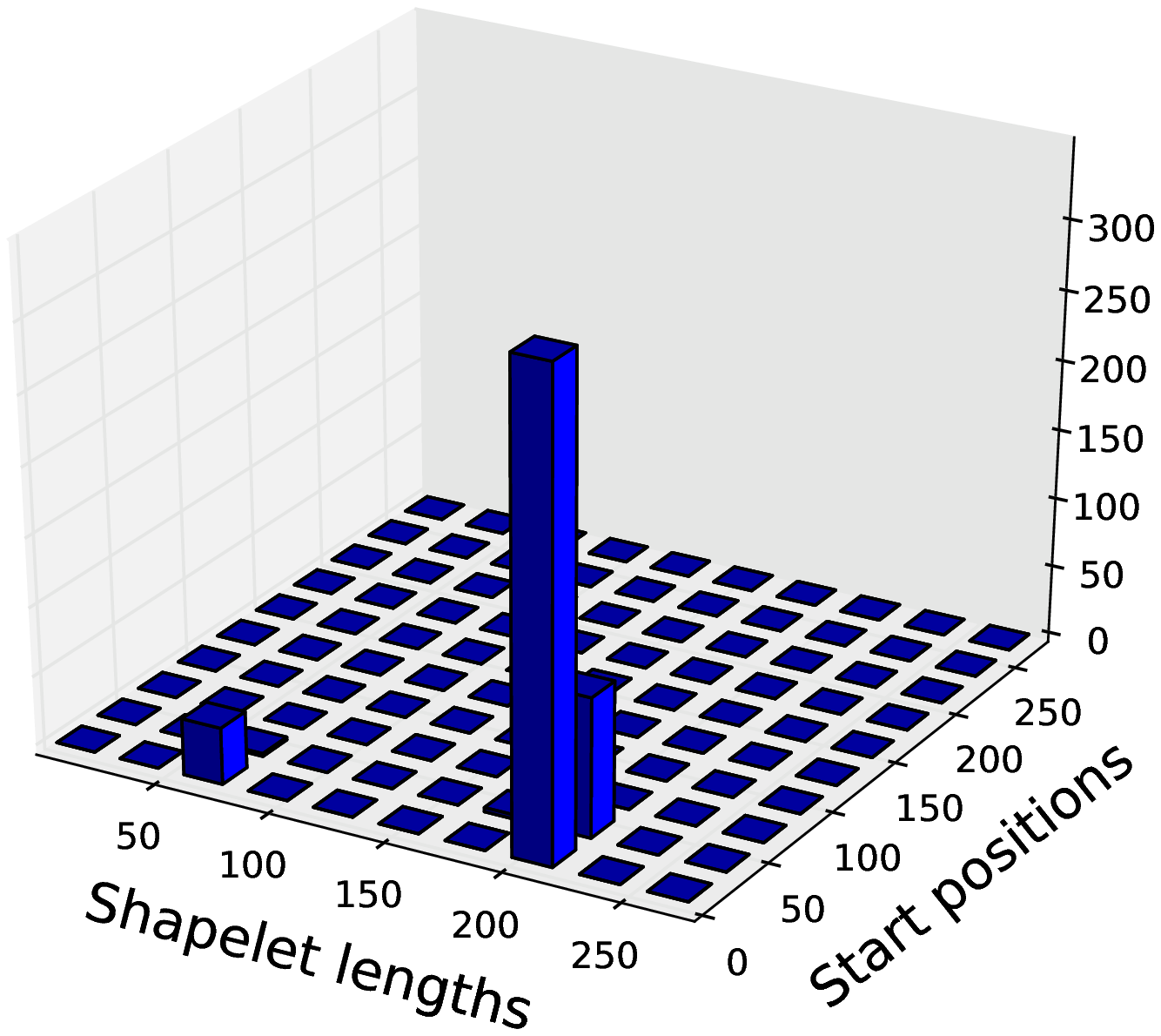,width=53mm}}
	\subfigure[controlCharts]{\label{fig:ControlCharts_hist}\epsfig{file=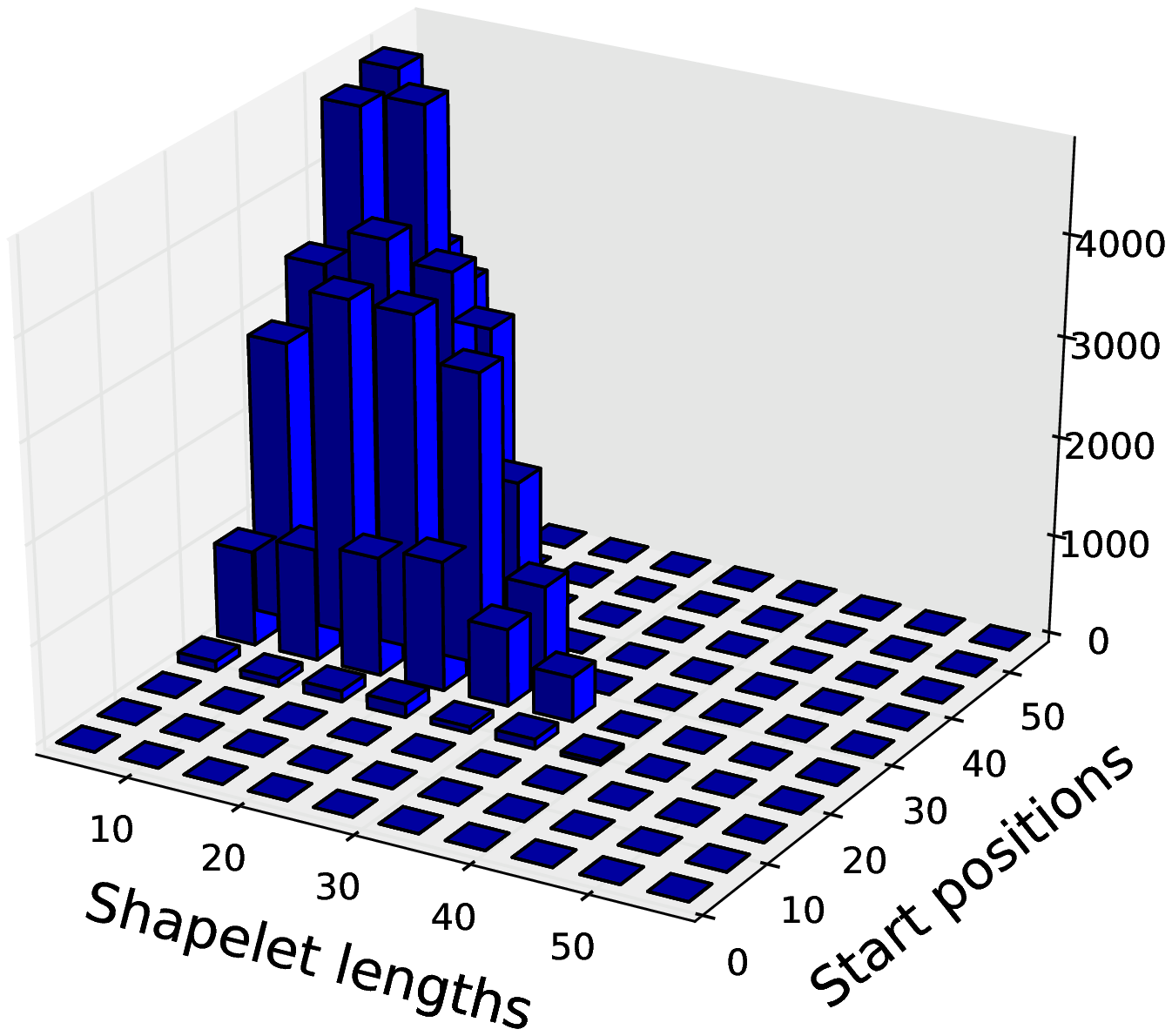,width=53mm}}
	\subfigure[ecgPatterns]{\label{fig:ecgPatterns_hist}\epsfig{file=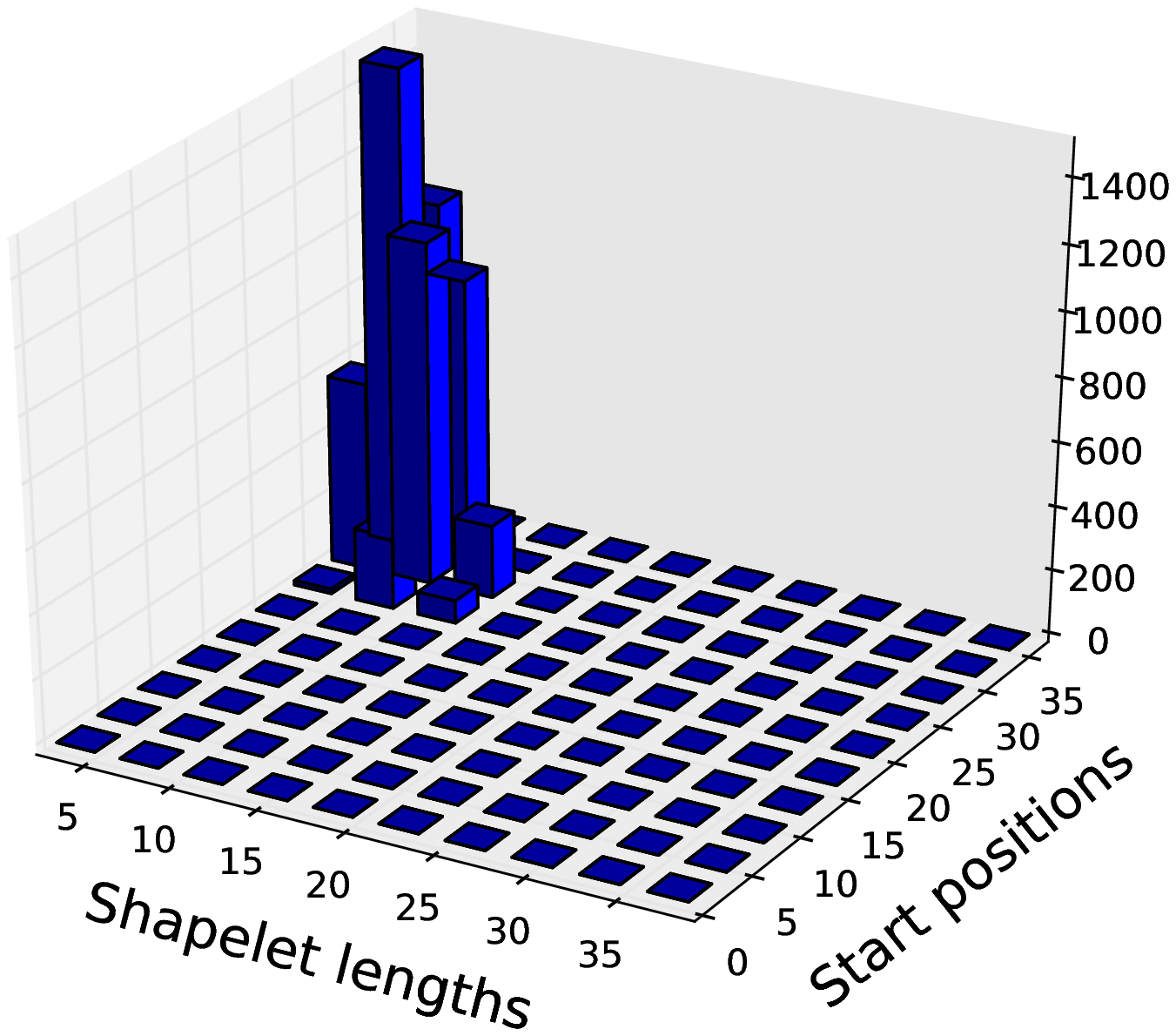,width=53mm}}
	\subfigure[mallat]{\label{fig:mallat_hist}\epsfig{file=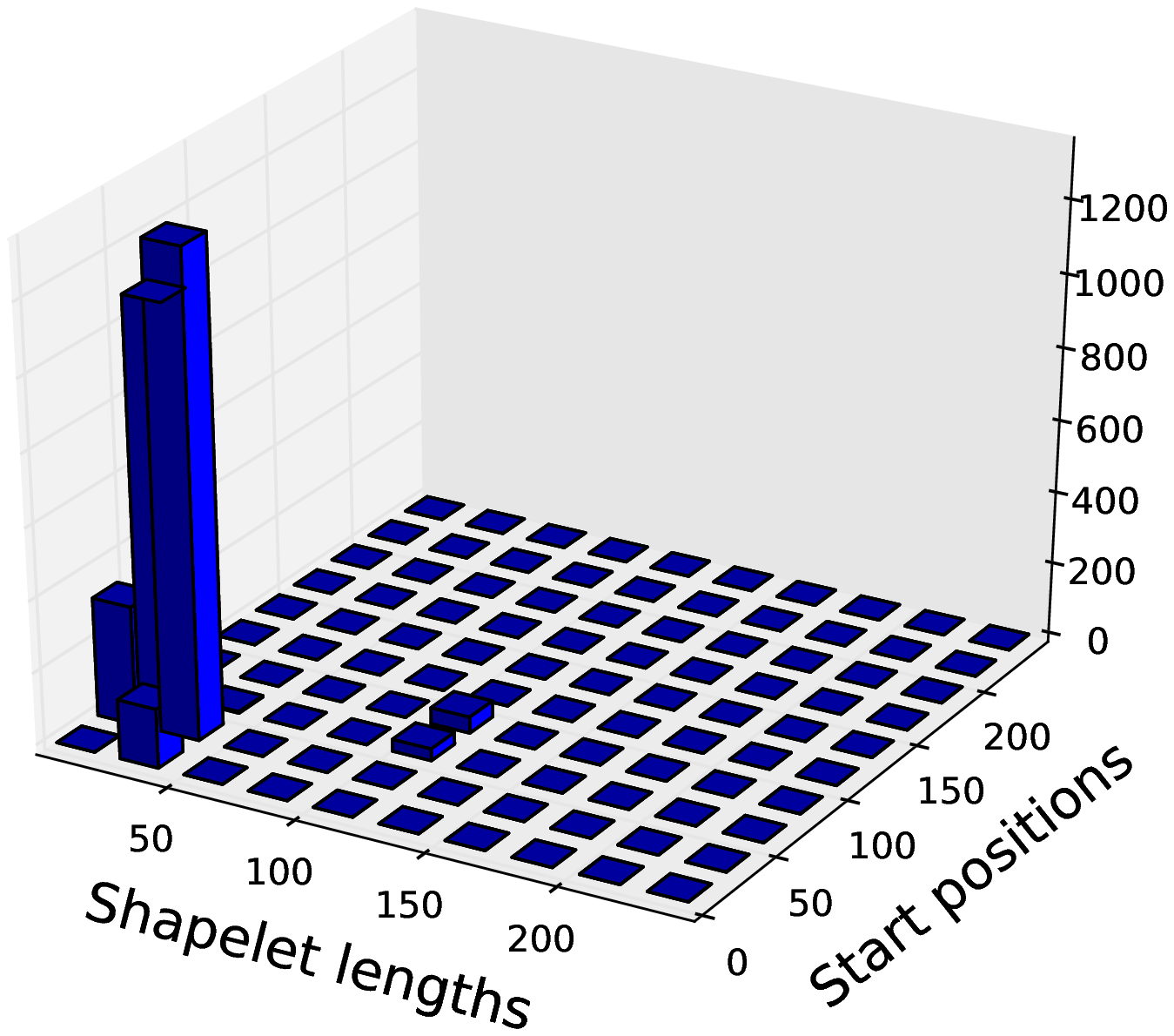,width=53mm}}
	\subfigure[shield]{\label{fig:shield_hist}\epsfig{file=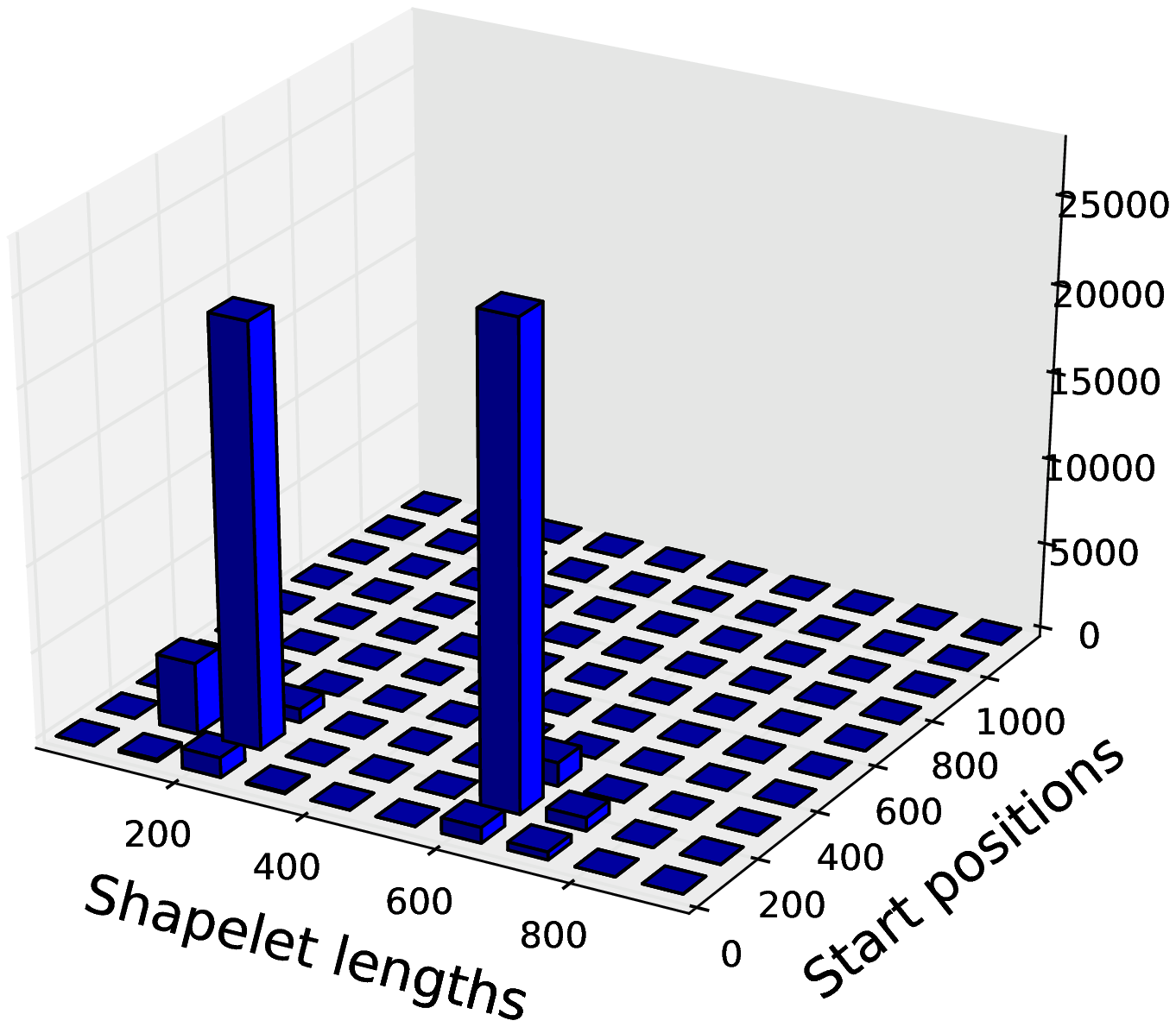,width=53mm}}
	\subfigure[wheat]{\label{fig:wheat_hist}\epsfig{file=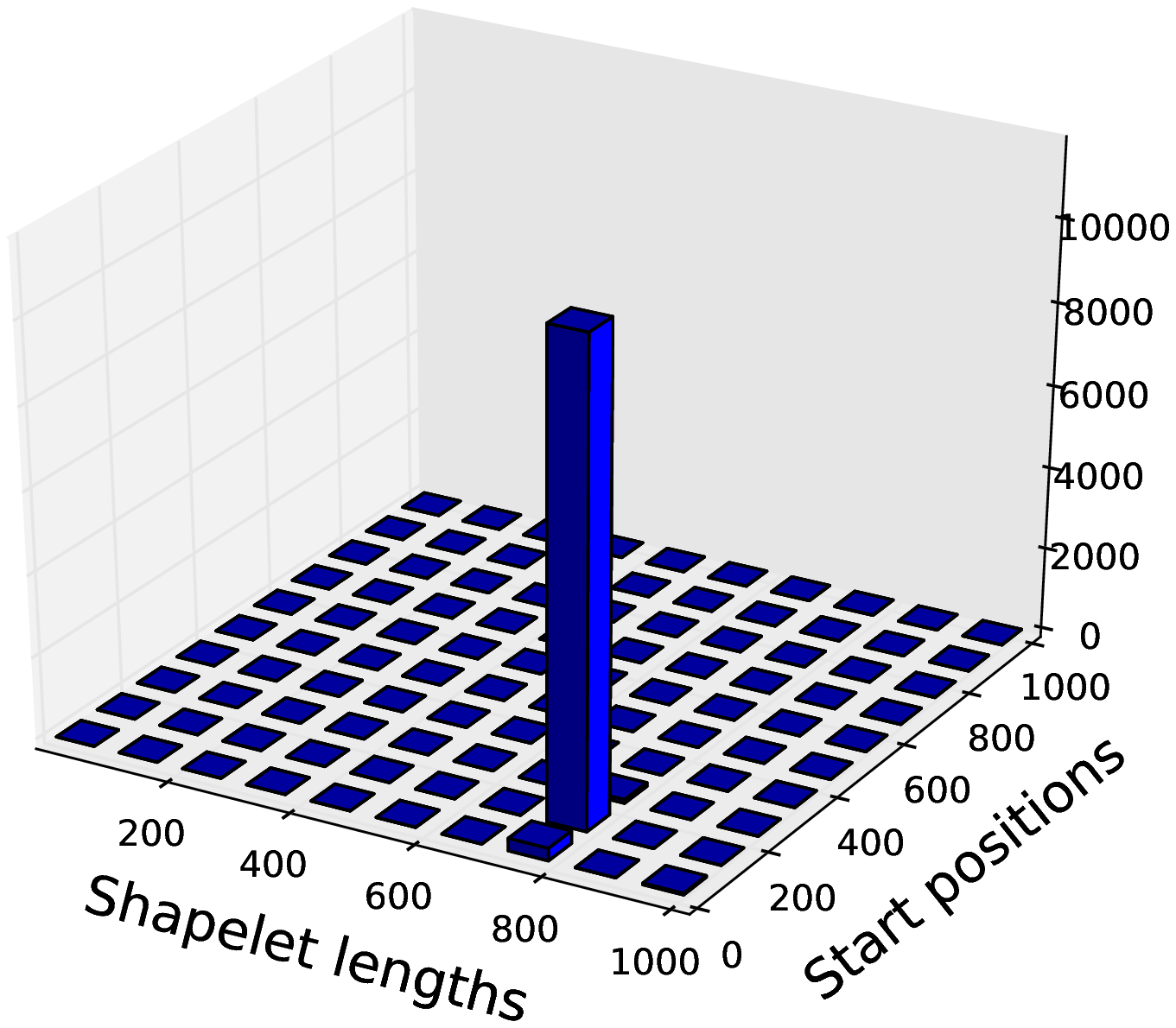,width=53mm}}
	\caption{Histograms of high-quality shapelet concentrations.
					 Axes are the lengths of the shapelets and the positions in the time series from which they were extracted}
	\label{fig:shConc}
\end{figure*}

%
%\begin{table}
%\centering
%\caption{Concentrations of high quality shapelets}
%\label{table:shConcentrations}
%\begin{tabular}{|l|l|l|}					 					\hline
%Dataset				&lengths	&start positions 	\\ \hline
%arrowhead 		&86-194		&	118-173				 	\\ \hline
%coffee				&18-38		& 65-79,					\\
%							&					&	198-213					\\ \hline
%controlCharts &19-60		&	0-39						\\ \hline
%ecgPatterns		&26-40		&	0-9							\\ \hline
%mallet				&22-86		&	11-114					\\ \hline
%shield				&60-175		&	225-390,				\\
%							&					&	747-927					\\ \hline
%wheat					&40-220		&	244,						\\
%							&					&	742-959					\\ \hline
%\end{tabular}
%\end{table}

Fig.~\ref{fig:shConc} shows the numbers of high-quality shapelets of different lengths and different offsets within the time series.
It is evident that high-quality shapelets are not evenly distributed through out the shapelet space;
rather they are concentrated in clusters. This fits with the main concept of shapelet classification, that there are specific areas inside time series which contain the classes ``fingerprint'', allowing differentiation between different classes.

As high-quality shapelets all appear in a particular area of the shapelet space,
scanning the entire space starting with the shortest length and continuing incrementally to shapelets of the longest length will, in most cases, require examining a considerable number of shapelets before reaching high-quality shapelets. This is the reason that both the \textit{simple} and the \textit{binary} orderings
do not attain high accuracy as early on as \textit{random} ordering.
This is also the reason that the enumeration used by the YK algorithm does not generally achieve high accuracy early on.
Although the YK algorithm creates a random permutation of all shapelets within each length,
it nevertheless enforces examination of shapelets from shortest to longest length.

As it is not known a priori where high-quality shapelets are clustered within the shapelets space, random sampling across different shapelet lengths and offsets is required.

\subsection{Performance of the SALSA-R Algorithm}
After establishing that \textit{random} is the best sampling order and gaining insights into why this is so,
we proceed to evaluate the SALSA-R algorithm. As we described in Section \ref{subsec:algorithm}, the SALSA-R algorithm uses the random evaluation order and outputs a classification model upon convergence to a high-accuracy model.

%Our algorithm self terminates after determining that the shapelet currently found to be best probably has IG and %margin values similar to the
%best shapelet in the training set.
%Based on the insights of sub-section \ref{subsec:randSuc} it is possible to claim that the shapelet will probably also  %be similar in shape
%to the best shapelet in the training set.

\subsubsection{Experimental setup}
We evaluated SALSA-R on all datasets presented in table~\ref{table:datasets}.
The value for $\epsilon$ was 0.01, i.e. our algorithm considers a shapelet as significantly better
if its IG or margin are better by a factor of at least $1\%$.
For the number of iterations \emph{NI}, after which SALSA-R terminates if no significant improvement in shapelet quality was observed, we tested two values: $10,000$ and $100,000$.
For each combination of $\epsilon$ and \emph{NI} and for each dataset, we executed our procedure $20$ times, due to the element of randomness.

\subsubsection{Experimental Results}
For each dataset, we calculated the average accuracy, number of shapelets examined and the running time.
The results are presented in table ~\ref{table:selfTerm}. For each measure we present the values obtained using $NI=10,000$, $NI=10,0000$ and those achieved by prior art.
By prior art we refer to both the YK algorithm and to a more optimized implementation described in \cite{mueen2011logical}. When we could, we compared with the faster procedure (available at \cite{logicalCode})
which can discard a shapelet by comparing it to previously examined shapelets, but due to large RAM consumption\footnote{We tried running the pruning algorithm on all of our datasets, but for mallat,
shield and wheat the implementation of \cite{logicalCode} exited abnormally after using 2GB of RAM, as it was compiled for 32-bit memory space.},
for datasets mallat, shield and wheat, we compared with the YK algorithm
(for mallat, the YK algorithm did not terminate after more than two weeks so we compared with the accuracy published in \cite{ye2011time}).
Next to each average accuracy, we recorded the standard deviation in brackets.

%The value of $\epsilon$ was chosen based on additional experiments, results of which are not presented here for lack %of space. While examining possible values for \emph{NI}, we noticed that, while the average accuracy doesn't change %much, for small values of \emph{NI}, the standard deviation is usually bigger, implying that using a larger value of %\emph{NI } (still orders of magnitude smaller than the size of the shapelets space) will probably return a better %model.
%The higher variance with smaller values of \emph{NI} seems arises because the number of common shapelets examined is %much
%smaller, leaving more room for big differences.
%Therefore it seems that specifying $\epsilon=0.01,\ \  NI=100,000$, if possible (for very large datasets even $NI=100,000$ will take too long),
%will produce a good model.

\begin{table*}
\centering
\caption{Performance of self terminating algorithm with $\epsilon=0.01$ and $NI=10,000,\ \  100,000$}
\label{table:selfTerm}
\begin{tabular}{|l||c|c|c||c|c|c||c|c|c|}					 					\hline
Dataset				&\multicolumn{3}{|c||}{Accuracy($\%$) (std)}	&\multicolumn{3}{|c||}{number of shapelets} & \multicolumn{3}{|c|}{time (sec)}\\ \hline
							&NI=10,000			&NI=100,000	&Prior Art				&NI=10,000	&NI=100,000	&Prior Art				 &NI=10,000	&NI=100,000	&Prior Art	\\ \hline
arrowhead 		&78.1	(4.5)			&78.4	(2.8)	&80.0							&1.8e4			&1.6e5			 &1.6e6						&28.7				&254.0			&2444.2			\\ \hline
coffee				&91.4	(5.3)			&95.0	(5.6)	&100.0						&1.6e4			&1.7e5			 &1.1e6						&10.4				&103.1			&492.7			\\ \hline
controlCharts &89.2	(2.5)			&87.1 (1.4)	&85.7							&1.2e4			&1.5e5			&3.4e5						 &9.5				&112.6			&1613.2			\\ \hline
ecgPatterns		&90.3	(3.1)			&85.0	(0.0)	&85.0							&1.1e4			&7.4e4	 		 &7.4e4						&2.2				&15.0				&90.0				\\ \hline
mallat				&99.0	(0.4)			&98.7 (0.7)	&98.8							&1.8e4			&1.8e5			 &1.0e7						&114.4 			&954.0			&$>$1.2e6		\\ \hline
shield				&87.3 (1.3)			&88.1	(1.3)	&89.1							&1.6e4			&1.6e5			 &1.7e7						&135.6			&994.3			&4.2e5			\\ \hline
wheat					&67.7 (2.2)			&69.2 (2.0)	&58.0							&1.6e4			&1.7e5			 &2.7e7						&214.1			&1773.6			&5.1e5			\\ \hline
\end{tabular}
\end{table*}

%\begin{table}
%\centering
%\caption{Concentrations of high quality shapelets}
%\label{table:shConcentrations}
%\begin{tabular}{|l|l|l|l|}					 					\hline
%Dataset				&lengths	&start positions 	& shapelets	\\ \hline
%arrowhead 		&86-194		&	118-173				 	& 1,407		\\ \hline
%coffee				&18-38		& 65-79,					& 496			\\
%							&					&	198-213					&					\\ \hline
%controlCharts &19-60		&	0-39						& 54,813	\\ \hline
%ecgPatterns		&26-40		&	0-9							& 5,554 	\\ \hline
%mallat				&22-86		&	11-114					& 2,996		\\ \hline
%shield				&60-175		&	225-390,				& 62,473	\\
%							&					&	747-927					& 				\\ \hline
%wheat					&40-220		&	244,						& 12,078	\\
%							&					&	742-959					&					\\ \hline
%\end{tabular}
%\end{table}

For all datasets except one, the accuracy we achieve is very close to that of prior art (within $2\%$),
and for three of the datasets our algorithm improves the accuracy obtained.
Also, the standard deviation is relatively small, indicating that the expected accuracy of any single model should be close to the average value.
The numbers of shapelets checked show that SALSA-R examines only a small fraction of the shapelets space
(except for ecgPatterns and controlCharts which are small datasets), in some cases only $1/1000$ of all shapelets,
while still attaining high accuracy.
Naturally, the number of shapelets examined is reflected in the time required to build a model,
as recorded in the last three columns of table ~\ref{table:selfTerm}.
For small datasets, our algorithm terminates (on average) after a few seconds, compared with previous algorithms which require a number of minutes.
For larger datasets, a good model is computed within a few minutes instead of a number of days.
This comparison establishes that for most datasets, the SALSA-R algorithm
outputs an accurate model using only a very small fraction of the shapelets required by the pruning method and requiring only a small fraction of the time. Moreover, the performance boost gained by SALSA-R grows quickly with the size of the dataset.

\section{Conclusions}
\label{sec:conclusions}
The focus of our work was acceleration of the time taken to build a model for shapelet-based classification.
To this end, we compared different orders for shaplet evaluation,
concluding that from all those examined, the \textit{random} order is best.
Our work establishes that the reason for this is that shapelets are not evenly distributed through out the entire shapelets space. Rather, they are concentrated in a tight cluster of shapelet lengths and locations in time series. Consequently, fast identification of high-quality shapelets requires quick sampling of the entire shapelets space.

We implemented SALSA-R, an algorithm that samples shaplets randomly and uniformly from the shapelets space and outputs a classification tree model upon observing that the quality of sampled shapelets stabilizes. Our evaluation shows that SALSA-R outputs an accurate model after evaluating only a small fraction of the number of shapelets required by prior art shapelet-based learning algorithms, requiring only a small fraction of their time. A novel observation stemming from our work is that considering too many shapelets may lead to a decrease in model accuracy, most probably because of overfitting.

%Our main conclusions are as follows:
%\begin{enumerate}
%	\item Examining the shapelets domain in random order achieves high accuracy even after considering a small fraction %of all shapelets.
%	\item Considering too many shapelets may lead to a decrease in accuracy. We suspect that this is due to %overfitting.
%\end{enumerate}

In the future we plan to attempt to further increase the accuracy and reduce the time-complexity of our algorithm. One direction is to try alternative sampling strategies. For instance, accuracy may be improved by initially performing random and uniform sampling of the shaplets space and then reverting to a more thorough search of specific areas in which a significant number of high-quality shaplets were found. The current measure for shaplet quality is its information gain and margin. Another avenue for future work is to investigate alternative shapelet quality criteria.

%\textbf{Future work} -
%\begin{inparaenum}[1)]
%\item The time required for examining each shapelet is long, as a shapelet needs to be compared with all possible %positions in
%every time series. Therefore our current implementation cannot cope with very large datasets.
%\item Our results also indicate that using IG and margin for choosing good shapelets may fail to detect the best %shapelet.
%Therefore additional methods for measuring a shapelets quality are required.
%\item Currently our random search focuses equally on the entire domain. It may be possible to improve accuracy, by %first scanning the entire domain,
%locating the area with good shapelets and then performing a more thorough search of the subdomain rich in good %shapelets.
%\end{inparaenum}
%major - anytime, random, curse
%minor- better understanding of correlation between measure and accuracy
%Discuss random sampling with repetition.
%Discuss using only a small randomly chosen subset of the subsequences
%Discuss the fact that IG does not increase monotonically.
%The algorithm still cannot be used for very large datasets. 

%ACKNOWLEDGMENTS are optional
%\section{Acknowledgments}
%We wish to thank A.Mueen and E.Keogh for providing the code for their algorithms,
%and for taking the time to answer our many questions.
%Thank Keogh for his code.
%Thank DT

%
% The following two commands are all you need in the
% initial runs of your .tex file to
% produce the bibliography for the citations in your paper.
\bibliographystyle{IEEEtran}
\bibliography{IEEEabrv,references}
\end{document}